\documentclass[9pt]{article}
\usepackage[left=1in, right=1in, top=1in]{geometry}

\usepackage[utf8]{inputenc}
\usepackage{hyperref}
\usepackage[
backend=biber,
style=numeric,
sorting=none
]{biblatex}
\usepackage{caption}
\usepackage{graphicx}
\usepackage[export]{adjustbox}
\graphicspath{ {figures/} }

\usepackage{authblk}

\usepackage{listings}

\usepackage{xcolor}

\colorlet{punct}{red!60!black}
\definecolor{background}{HTML}{EEEEEE}
\definecolor{delim}{RGB}{20,105,176}
\colorlet{numb}{magenta!60!black}

\lstdefinelanguage{json}{
	basicstyle=\tiny\ttfamily,
	numbers=left,
	numberstyle=\tiny,
	stepnumber=1,
	numbersep=8pt,
	showstringspaces=false,
	breaklines=true,
	frame=lines,
	backgroundcolor=\color{background},
	literate=
	*{0}{{{\color{numb}0}}}{1}
	{1}{{{\color{numb}1}}}{1}
	{2}{{{\color{numb}2}}}{1}
	{3}{{{\color{numb}3}}}{1}
	{4}{{{\color{numb}4}}}{1}
	{5}{{{\color{numb}5}}}{1}
	{6}{{{\color{numb}6}}}{1}
	{7}{{{\color{numb}7}}}{1}
	{8}{{{\color{numb}8}}}{1}
	{9}{{{\color{numb}9}}}{1}
	{:}{{{\color{punct}{:}}}}{1}
	{,}{{{\color{punct}{,}}}}{1}
	{\{}{{{\color{delim}{\{}}}}{1}
	{\}}{{{\color{delim}{\}}}}}{1}
	{[}{{{\color{delim}{[}}}}{1}
	{]}{{{\color{delim}{]}}}}{1},
}

\usepackage[toc,page,titletoc,title]{appendix}

\title{\textbf{Evolutionary Cell Aided Design for Neural Network Architectures}}
		
\author[1,3]{Philip Colangelo\thanks{philip.colangelo@intel.com}}
\author[2]{Oren Segal\thanks{oren.segal@hofstra.edu}}
\author[2]{Alexander Speicher\thanks{aspeicher1@pride.hofstra.edu}}
\author[1]{Martin Margala\thanks{Martin\_Margala@uml.edu}}
\affil[1]{Department of Computer Engineering, University of Massachusetts Lowell}
\affil[2]{Department of Computer Science, Hofstra University}
\affil[3]{Intel PSG}

\begin{document}

\date{}

\maketitle

\begin{abstract}
Mathematical theory shows us that multilayer feedforward Artificial Neural Networks(ANNs) are universal function approximators, capable of approximating any measurable function to any desired degree of accuracy. In practice designing practical and efficient neural network architectures require significant effort and expertise. We present a novel software framework called Evolutionary Cell Aided Design(ECAD) meant to aid in the exploration and design of efficient Neural Network Architectures(NNAs) for reconfigurable hardware. Given a general neural network structure and a set of constraints and fitness functions, the framework will explore both the space of possible NNA and the space of possible hardware designs, using evolutionary algorithms, and attempt to find the fittest co-design solutions according to a predefined set of goals. We test the framework on an image classification task and use the MNIST data set of hand written digits with an Intel Arria 10 GX 1150 device as our target platform. We design and implement a modular and scalable 2D systolic array with enhancements for machine learning that can be used by the framework for the hardware search space. Our results demonstrate the ability to pair neural network design and hardware development together using an evolutionary algorithm and removing traditional human-in-the-loop development tasks. By running various experiments of the fittest solutions for neural network and hardware searches, we demonstrate the full end-to-end capabilities of the ECAD framework.

\end{abstract}

\section{Introduction}

The difficulty in designing performant NNAs has brought a recent surge in interest in auto design of NNAs. The focus of the existing body of research has been on optimizing NNA design for accuracy \cite{liu2017hierarchical}\cite{real2017large}\cite{real2018regularized}. Optimizing NNAs is typically a difficult process in part because of the vast number of hyperparameter combinations that exist, and in cases where a combination is not optimal, performance will suffer. In fact, many deep learning frameworks such as TensorFlow \cite{abadi2016tensorflow} and Keras \cite{gulli2017deep} offer support for hyperparameter tuning, but these are typically Bayesian optimizations that treat the ANN as a black-box and are used for the neural network training process. Research shows that the parameters of a network can directly influence the accuracy, throughput, and energy consumption of that model in deployment \cite{canziani2016analysis}. 

Once an accurate NNA has been found, 
the next step is to try to fit it into existing hardware i.e. a CPU, GPU, or a custom built but general purpose neural network hardware device such as a TPU \cite{jouppi2018motivation}. 
None of these hardware solutions offer network specific specialization.
The gap between the two optimizations is where ECAD comes in, it allows to search for an optimal hardware/NNA co-design by exploring the design space on the NNA and the hardware side and allows to implement a custom hardware solution for a specific NNA model using reconfigurable hardware.

\section{Related Work}

In the past several years we have seen great strides made in the performance of ML algorithms on complex tasks using deep neural networks \cite{krizhevsky2012imagenet}\cite{krizhevsky2009learning}. 
Manually designing state of the art deep neural network architectures for ML requires significant amount of time and labor \cite{real2017large}\cite{elsken2018neural}.
Automating NNA search has been an ongoing effort for the past few decades but is becoming a focus of the NNA research community because of the difficulty in designing deep networks which are ever growing in complexity\cite{real2017large}\cite{real2018regularized}\cite{elsken2018neural}. 
Automatic Artificial Neural Network Architectures Search (NAS) can be conducted using different strategies such as random search, evolutionary algorithms, Reinforcement Learning (RL), Bayesian optimization, 
and gradient-based methods \cite{elsken2018neural}.
Using Evolutionary Algorithms (EAs) to search for performant architectures has been investigated extensively \cite{miller1989designing}\cite{stanley2002evolving} over the years. 
Some recent results indicate that evolutionary algorithms offer better results than random search and reinforcement learning \cite{real2018regularized}
Recently, there has been growing interest in NAS for deep neural networks that specialize in image recognition \cite{liu2017hierarchical}\cite{real2017large}\cite{Zoph2018LearningTA}. 

As deep and complex neural networks became increasingly popular and with the realization that existing hardware architectures are not specifically optimized for such computation, new forms of specialized architectures have been proposed and designed \cite{jouppi2018motivation} to help increase performance and energy efficiency. Designing such static new architectures can be prohibitively expensive and risky since the field of neural computing is evolving so rapidly.
Optimizing hardware for neural networks is a research topic that is constantly evolving and correlating with the ever-changing network structures that are being developed. Recent publications have shown new architectures carving their niche in deep learning by offering unique methods for accelerating the workloads of neural network applications \cite{DBLP:journals/corr/AydonatOCLC17}\cite{Suda:2016:TOF:2847263.2847276}\cite{DBLP:conf/icassp/ParkS16}. Specifically, these reconfigurable architectures are a popular platform for both research and deployment due to their ability to change their fabric routing and resource structures to fit various workloads and optimizations by leveraging the resiliency of neural networks through low-numeric precision \cite{DBLP:journals/corr/abs-1806-11547}\cite{DBLP:journals/corr/UmurogluFGBLJV16} and sparsity \cite{DBLP:conf/icassp/VenkateshNM17}\cite{8280155}. Other accelerator designs use reconfigurable fabrics to change the logic routing to preprocess data and offload to a specialized ASIC \cite{Nurvitadhi:2018:IDA:3174243.3174966}. 

Multiple tool flows exist for optimizing fixed NNA designs for reconfigurable hardware (FPGAs). The majority of available tool flows target image recognition tasks. A recent survey on available tool flows is available here \cite{venieris2018toolflows}. 

The body of work on NAS concentrate on accuracy as the main measure of performance, though optimizing for NAS can lead to more simplified NNA that could in turn simplify and optimize hardware designs \cite{real2018regularized}\cite{elsken2018neural}. On the other hand, optimizing for hardware performance parameters (latency/throughput/power) is normally done on an existing NNA design and there is no attempt to modify the NNA (layers/neurons etc.)\cite{venieris2018toolflows}. 

Combining NAS and hardware optimizations could potentially close the loop between design and implementation of NNAs\cite{venieris2018toolflows}.      

To the best of our knowledge ECAD is the first framework capable of conducting NAS and hardware co-optimization. It is capable of working on both the NNA level (neurons/layers etc.) and the hardware level (LUTS/DSPs etc.) at the same time i.e. given a general NNA structure it will evolve and search both spaces (NNA/hardware) in tandem.

\section{ECAD Software }
ECAD is intended to create a NNA that is optimized towards specific design goals.
At the heart of the software side, lies an evolutionary algorithm and a vector of fitness functions. Fitness functions currently include measurements of accuracy, speed, energy efficiency, and throughput. ECAD allows to select the importance(weight) given to each of the fitness functions and by doing so guide an evolutionary process towards the required fitness goals. The result is a neural network design optimized towards the goals specified in the fitness functions.

\subsection{ECAD Software Flow}
\begin{figure}
	\center
	\includegraphics[scale=0.50]{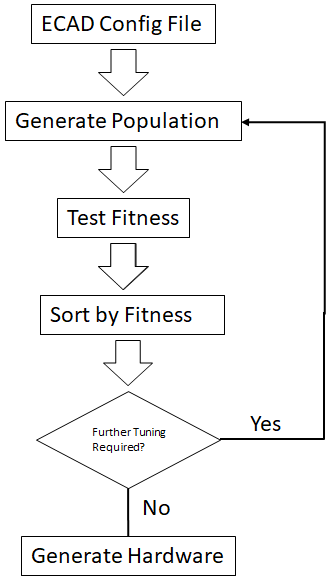} 			
	\caption{ECAD flow.}
	\label{fig:ecad_flow}	
\end{figure}

Figure \ref{fig:ecad_flow} shows an overview of the ECAD flow. The next sections provide detail for each of the stages in the flow.
\subsubsection{The ECAD Configuration File}
The process starts with a user generated description of the desired neural network. It includes a description of the structure, constraints, fitness goals and weight of each goal. Those values are stored in the ECAD configuration file in a textual JavaScript Object Notation (JSON) format. An example of the configuration file can be seen in 
Listing \ref{lst:ecad_cfg_lst}.
Lines 11 to 16 define population values such as initial and maximum population size, mutation rate etc. Lines 17 to 22 specify the worker types that will be used to evaluate our goals and their parameters. lines 33 to 90 declare cell types and their parameters such as range of legal values, mutation rate and user defined functions to be called upon a change event such as a mutation. Lines 99 to 109 define the hardware we wish to target.
lines 111 to 117 declare a cell array that will hold the general structure of the network we would like to explore through the evolutionary process.

\subsubsection{Population Generation} \label{pop_gen_subsec}

Initially the system will create a population of neural networks using the base design specified in the configuration file. The population initial size, maximum size and change rate  are all controlled using the configuration parameters. Each auto generated network instance will be mutated and different from the original base design.

As the evolutionary process progresses and once a sufficient number of networks are evaluated according to all fitness parameters, the process of population generation will repeat itself continuously except that the mutations will be based on the most fit individuals in the population, selected according to their fitness scores (see steady-state model in \cite{goldberg1991comparative}). 

\subsubsection{Testing Population Fitness}

In the next stage each NN instance will be sent to one or more fitness evaluators or workers in ECAD terminology. The workers are designed as independent processes and can be distributed across a cluster of computers. They are orchestrated using a Master/Worker parallel computation model \cite{rauber2013parallel} running on top of MPI \cite{gabriel2004open}.

Each Worker sits in a separate process, inherits from a common C++ Worker class, and implements a common software interface. 
The system is built to be flexible and allow adding different types of workers easily. 
The implementation details for each worker can be completely different.

We currently have three types of workers implemented:

\begin{enumerate}
	\item \textit{Simulation Worker} capable of simulating a NN design and return accuracy and timing results 
	\item \textit{Physical Worker} capable of synthesizing a NN design and return hardware synthesis results 
	\item \textit{HWDB Worker} capable of accurately estimating NN synthesis results and return estimated hardware synthesis results  
\end{enumerate}

Once a worker's fitness evaluation is complete it will return the results to the master/server process. The master process will collect the results from all the workers that evaluated each NN and apply a combined score to each NN. Scores are combined using a unique Id that is assigned to each generated NN instance when it is initially created.

\subsubsection{Sorting Population by Fitness}

In the next stage the ECAD system will sort the NN population according to the score each NN received. The top NN instances will be selected for mutation. 

The mutation process will take the top NN performers and mutate the cell/layer fields randomly according to constraints specified in the configuration file. For example, the dense layer could specify a range of allowed number of neurons etc.

The mutated NN will be introduced to the population as new instances (\ref{pop_gen_subsec}). Note that if we limit the population size in the configuration file and the addition of the new instances will cause an overflow, the worst performers will be removed from the population in this step as well.

\par 
We now have a new population containing a mix of old and new instances. New instances will be sent to evaluation and the evaluation process will repeat until the end condition is met. 

The end condition could be the number of generations the simulation flow is requested to run or a desired accuracy.   

During the evaluation process, ECAD outputs the top networks and their scores to a custom database (EcadDB). At the end of the simulation this database is examined, and the top networks can be extracted for further investigation and full hardware synthesis. Note that even though it is possible to run full synthesis during the ECAD flow we use the HW resource estimator option because of resource and time constraints. A full hardware compile can take several hours per network instance and so it is reserved for top network candidates that deserve further investigation. 

\subsection{From Abstract Network Description to Concrete Design}
\begin{figure}
	\center
	\includegraphics[scale=0.50]{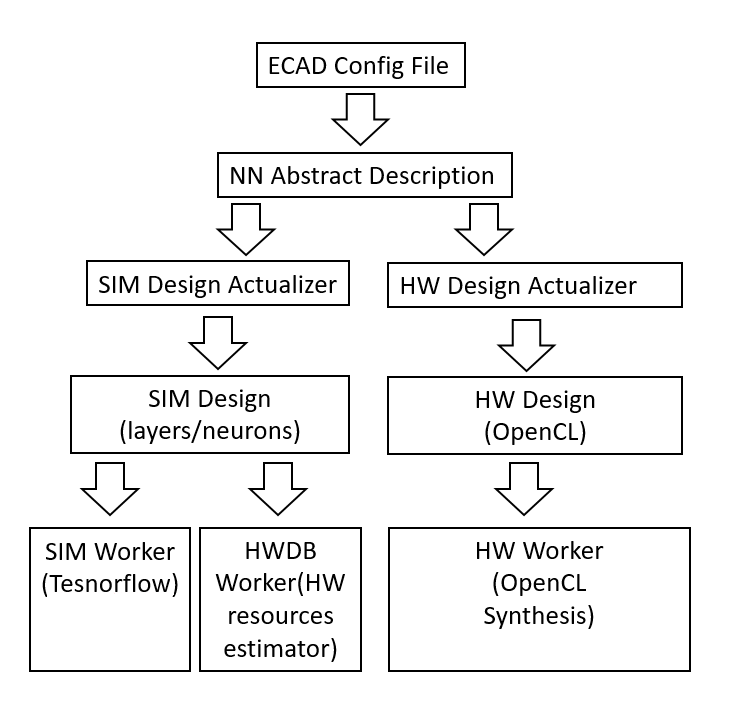} 			
	\caption{ECAD description to design Transformations.}
	\label{fig:ecad_transformations}	
\end{figure}

The process by which an abstract network description is transformed to a hardware or simulation design is depicted in Figure \ref{fig:ecad_transformations}. The software layer is built to allow the transformation to be general. For each type of transformation we wish to make, we design a C++ object of type \textit{Actualizer} (or a writer) which takes the abstract design and produces a model that we can test in software or hardware.

For hardware design generation we opted to use high-level OpenCL over low-level Hardware Description Language(HDL) since it allows auto generated designs to be more easily examined and maintained by human engineers. But since the system is flexible we can always move away from high-level to the HDL level if the need arises. Software level simulation of NN designs is used for testing accuracy, verifying results and estimating hardware resource usage without going through the costly hardware compilation process. For software level accuracy and validation we use a neural network simulator.

\subsection{Neural Network Simulator}

The Neural Network Simulator (NeuralNetSim) is responsible for testing and verifying the ECAD neural network architectures. 
We chose TensorFlow\cite{abadi2016tensorflow} as the machine learning simulation framework as it supports both CPU and GPU training for deep neural networks, it enjoys continuous support and updates in addition to a high-level python API. This allows for quick trouble shooting as well as easily adding new features, such as different types of layers or activation functions to the simulator.
The NeuralNetSim consists of three classes:  
\begin{enumerate}
	\item \textit{ECAD Reader} servers as an interface to the evolutionary framework. It's main responsibility is accepting inputs such as files and training arguments as well as returning data and reports after training has been completed. It extracts the network info from the ECAD file and passes it to the TF Model Builder.
	\item \textit{TF Model Builder} holds the actual TensorFlow graph and all other TensorFlow variables. It is responsible for dynamically creating the graph based on the info passed by the reader. It loads and handles the training and testing data. Finally it also holds the training functions which are called from the ECAD reader. 
	\item \textit{TF Functions} is a collection of TensorFlow API functions that are called by the TF Model Builder when building the graph. New TensorFlow functions can be added here to expand support for other types of networks, optimizers, activation functions, etc. 
\end{enumerate}

\subsubsection{Neural Network Simulation Flow}
The ECAD reader is  a flexible script responsible for accepting an ECAD file as well as several other arguments that affect the training of the neural network;
\begin{enumerate}
	\item \textit{ECAD file} containing the networks architecture.  
	\item \textit{Destination Directory} used to return the report file, weights and biases.
	\item \textit{Epochs}  dictates how many times the training set is used for training.
	\item \textit{Batch size} sets the amount of samples that are passed into the network at a time during training.
	\item \textit{SaveWB} is an optimal argument that tells the simulator to export the weights and biases after training.
	\item \textit{verboseTF} displays potential warning or deprecation messages when the graph is created. 
\end{enumerate}

In the first stage, the ECAD neural network architecture is converted into a TensorFlow graph to be trained. The ECAD network consists of cells which either represent the input, hidden layers, activation functions, or output which are encapsulated in a cell array. This cell array is sequentially traversed and the graph is dynamically generated. Whenever a new hidden layer is created a reference to the layers weights and biases is kept, which can later be used for retrieval of the values after training. The data type for the neural network is currently float32 which is used for both the input data, the weights and the biases as well. 

\par After the full network graph is created the cost and optimizer are instantiated. Currently the softmax with cross entropy function is used for the MNIST \cite{MNISTDB} classification problem and the utilized optimizer is the AdamOptimizer\cite{kingma2014adam} which is set to a learning rate of 0.001. None of these functions are final and as the project expands more cost functions and optimizers can be added.  

\par The final step, before training can begin, is getting the training and testing data. In this case the MNIST data is imported using tf.keras.datasets
and the input data is converted to float32 numpy  arrays. All elements in the inputs are divided by 255 to normalize the values between 0 and 1. Since ECAD only supports MLP networks, as of now, the input samples are reshaped to arrays of 784 elements which is a flattened representation of  the 28x28 images of the MNIST data-set. The training and testing labels are converted to one hot arrays so that they can be used with the TensorFlows softmax with cross entropy function. 

\par After all related TensorFlow variables are created a TensorFlow session is instantiated and the training can begin. 
The training data is batched and fed to the network and optimizer for the defined amount of epochs. The optimizer will reduce the loss calculated by the given cost function for each batch. After each full pass of the training data, the epochs accuracy is noted and the final accuracy is returned to the ECAD reader. After the training has been completed the reader will create a report JSON file storing the networks name/id,  accuracy, the number of epochs, the full training time, and the batch size. 
The file is stored in the destination directory and is utilized by the evolutionary algorithm to pick the most optimal networks for further development. 
\par If the saveWB argument is set, the ECAD reader will call the references to the weights and biases and extract them as float32 arrays. These are then converted to binary files and stored in the destination 
directory. For a MLP network two files per layer are created, one weights and one biases file, and each are named after the cell from the ecad file for easy reference. These files can be utilized for testing the FPGA designs to further check if the actual design has benefits on a hardware level. To make the files more usable, each file contains 4 integers at the beginning which describe the dimensions of the stored data.

\section{Hardware Design}
Analysis of deep neural networks (DNNs) shows that while state-of-the-art networks are becoming more accurate over time, it often comes at the cost of increased parameter size leading to higher computational complexity and/or power consumption. The authors of \cite{canziani2016analysis} show the correlations between the number of operations required for inference, the time taken to classify an image or batches of images, the power consumption of various hardware, and the resulting top-1 accuracy of current top performing models. It is evident that each neural network has a unique structure that may or may not fit in a certain system, i.e., there are trade-offs related to system constraints like power consumption, latency, classification accuracy, or throughput. Benchmarks that are run to arrive at these correlations are typically executed on instruction set based architectures like CPU or GPU. Similar instructions being called for each network allows for a nice baseline, however, the intended solution space we are interested in also includes re-configurable hardware which provides a unique pipeline system capable of molding to a networks unique structure.

\par Field Programmable Gate Arrays (FPGAs) provide a ``sea of logic'' that can be programmed and reconfigured to create unique circuits by routing together various primitive building blocks like those found in Intel\textsuperscript{\textregistered}'s Arria 10 \cite{Arria10} FPGA. Intel's Arria 10 FPGA provides variable precision DSP blocks that can be configured for either integer based or single precision floating-point multiply and add operations, 8-input fracturable look-up table based ALMs for implementing various logic functions, and embedded M20K memory blocks. Leveraging the flexibility of the FPGA, we can find unique solutions for all machine learning models. 

\par
\begin{figure}
	\center
	\includegraphics[scale=0.40]{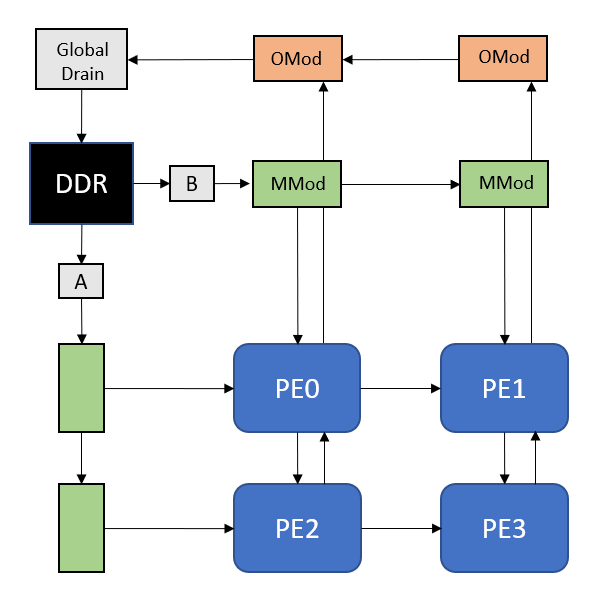} 			
	\caption{2D systolic array hardware architecture used in the design space exploration.}
	\label{fig:systolic_array}	
\end{figure}

 The evolutionary algorithm will guide a search for optimal FPGA hardware designs given a specific machine learning problem and optimization goals. Accomplishing this requires the evolutionary algorithm to have hooks into the FPGA solution space. Due to the nature of most DNN operations being vector and matrix based, we chose a 2-dimensional systolic array of processing elements (PEs) as shown in Figure \ref{fig:systolic_array}. The evolutionary algorithm has the capability to change the hardware configurations such as the array structures height, width, and PEs so that it may produce a new design for each permutation.

\par Every unique hardware configuration or permutation the evolutionary algorithm finds will be judged by a fitness function. Fitness functions are created to give a score so that future generations converge toward a more optimal design. Scores can be based on anything from power efficiency, throughput, or even logic utilization. Many solutions exist that would be functionally equivalent but have very different hardware. Some solutions will be computationally faster and others slower yet use less power. The ability to search for various levels of fitness such as performance, power, or cost allows a single design space to satisfy the needs of different vertical markets.  

\par Giving the evolutionary algorithm hooks into the hardware design space means providing a way for the evolutionary algorithm to modify the hardware description. Traditional hardware design done through a hardware description language such as Verilog allows for complete control over the hardware but does not provide a modular, software-like paradigm that would make this process straightforward. We address this issue by using OpenCL \cite{OpenCL} which is a high-level, C99-based programming language that can be used with Intel's FPGA SDK for OpenCL \cite{IntelOpenCL} to target Intel FPGA devices.

\subsection{2D Systolic Array Implementation} \label{2D Systolic Array Implementation}
\par Systolic arrays \cite{kung1979systolic} are a great fit for FPGA because of their pipelined data flow and memory bandwidth tuning capabilities. They can be scaled across one or many devices allowing for efficient data and model parallel solutions where each array has the capability to be uniquely configured. They are also modular. Each PE is designed to do a portion of the work, typically computing a partial result as a dot-product, and can be replaced with different processing types, e.g., low-bit, dense, or sparse.

Figure \ref{fig:systolic_array} shows the high-level architecture for our systolic array implementation which includes a couple enhancements for machine learning applications. Full descriptions of each module of our design will be discussed in the following sections. 

\subsubsection{Matrix Blocking and DDR Storage Considerations}
\begin{figure}
	\center
	\includegraphics[scale=0.40]{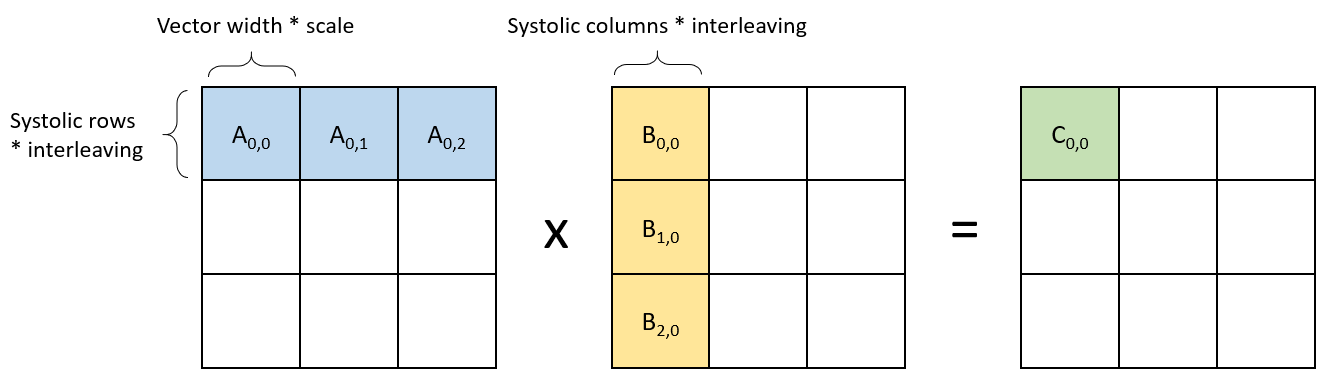} 			
	\caption{Matrix blocking example.}
	\label{fig:matrix_blocking}	
\end{figure}

Due to the nature of systolic arrays (also referred to as a grid in this text), input matrices are blocked against the spatial configurations of the architecture as shown in Figure \ref{fig:matrix_blocking}. Matrix A blocks have a height that is defined by the number of rows in the grid multiplied by an interleaving factor. The height of a matrix B block is the same as the width of a matrix A block and is defined by the product of vector width and scaling factor. The width of matrix B is defined as the number of columns in the grid multiplied by an interleaving factor. Interleaving is a parameter that is adjusted for balancing data reuse to ease bandwidth constraints. The larger the interleaving factor, the more data reuse and less bandwidth that is required to feed the PEs, but as the block size grows it becomes less efficient for mapping to smaller matrix sizes. Scaling factor is a parameter that is used to enable more efficient global memory access. We treat global memory as a 512-bit cache line. Every trip to global memory reads in a vector width amount of data and if the vector width is less than 512-bits then this can lead to sub-optimal bandwidth utilization. Scale will tune the block width so that each read to global memory is closer to 512-bits and more efficient. Blocks are stored contiguously in DDR memory to ease the global memory access patterns. Because of this, we transpose matrix B on the host so that sequential memory accesses traverse its rows.

\subsubsection{Loaders}
\par Depicted as A and B in Figure \ref{fig:systolic_array}, loaders are responsible for reading blocks in from global memory and sequencing the right data at the right time to the chain of memory modules so that each dot product computation is doing work towards the correct output matrix block. Part of this sequencing includes notifying the memory modules when the last block in a row or column has been sent, e.g. in Figure \ref{fig:matrix_blocking} block \(A_{0,2}\) and \(B_{2,0}\). Further, the loaders need to keep data flowing long enough to drain the last output matrix blocks back to global memory, so we incorporated a flush sequence that acts as a way to both re-initialize all accumulators and caches back to zero while allowing enough cycles to write all output data. During this sequence, instead of reading from global memory, the loaders simply send zero's along the memory modules. 

\subsubsection{Memory Modules}
\par Memory modules (MMods) are nothing more than a daisy chain of smart double buffers that read in the next block of data from the loader modules into a local cache while writing the current block to the PEs. MMods are chained along both the row and column dimension with the outer most module connected to a loader. Following Figure \ref{fig:mmod}, an MMod is made up of an input router whose job is to first direct a block of input to the write select demux before switching over to sending data down to its neighbor MMod. Once a cache such as Mem0 is full, the buff\_sel select will update so that new blocks are written to one memory while the read mux takes data from the other. Both memories are arbitrated in such a way that no read or write contention exists. All non-select lines depicted in  Figure \ref{fig:mmod} have a width that supports a vector worth of data. 

\begin{figure}
	\center
	\includegraphics[scale=0.55]{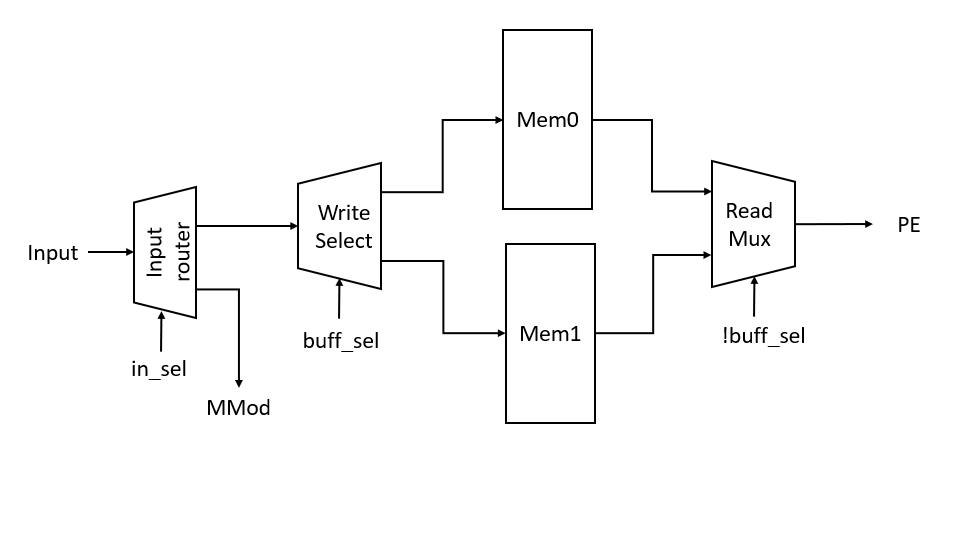} 			
	\caption{Memory module internals.}
	\label{fig:mmod}
\end{figure}

\subsubsection{Processing Elements} \label{pe}
\begin{figure}
	\center
	\includegraphics[scale=0.40]{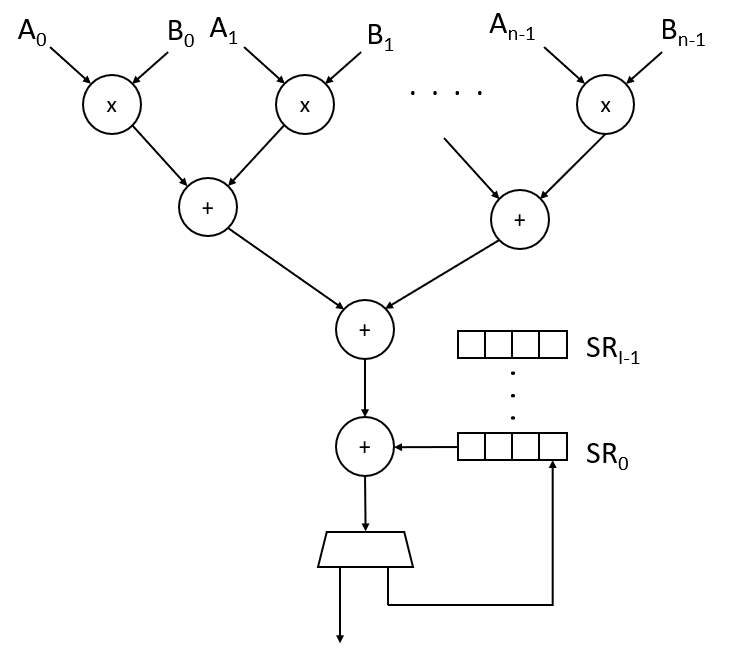} 			
	\caption{PE internals.}
	\label{fig:pe_internals}
\end{figure}

Each PE is responsible for computing a dot-product. Peripheral PEs (PE0, PE1, PE2 as seen in Figure \ref{fig:systolic_array}) get one input from a memory module (MMod) and the other from a neighbor except for the very first PE (PE0) who receives both inputs from MMods. Inner PEs (PE3) receive their input from both neighbor PEs. Each connection to a PE (except for the OMod connections that will be covered shortly) carries several data elements equal to the vectorization parameter of the array, or in other words, the width of the dot product. Figure \ref{fig:pe_internals} shows the internal workings of a PE. Each multiplier and adder in our design computes on 32-bit single-precision floating point data. The width of the dot product is depicted as n in the diagram. We chose a reduction tree strategy to make effective use of the DSP blocks and allow for deep pipelining of the design. PEs also include a small cache shown as shift registers (SR) in Figure \ref{fig:pe_internals}. The size of the shift registers is based on the interleaving factor (shown as I) Every cycle, a new vector enters the tree and is accumulated along with a previous value that is stored in one of the shift registers. The output from this accumulation is then routed to either the output or to the back of the shift register. The demux selector is based on a counter that keeps track of how many partial sums have been computed which signals when a result is ready. Once the counter rolls over, a drain sequence begins by routing the accumulated result out and starting a new output block sequence. 

\subsubsection{Output Modules and Global Drain}
\par
\begin{figure}
	\center
	\includegraphics[scale=0.40]{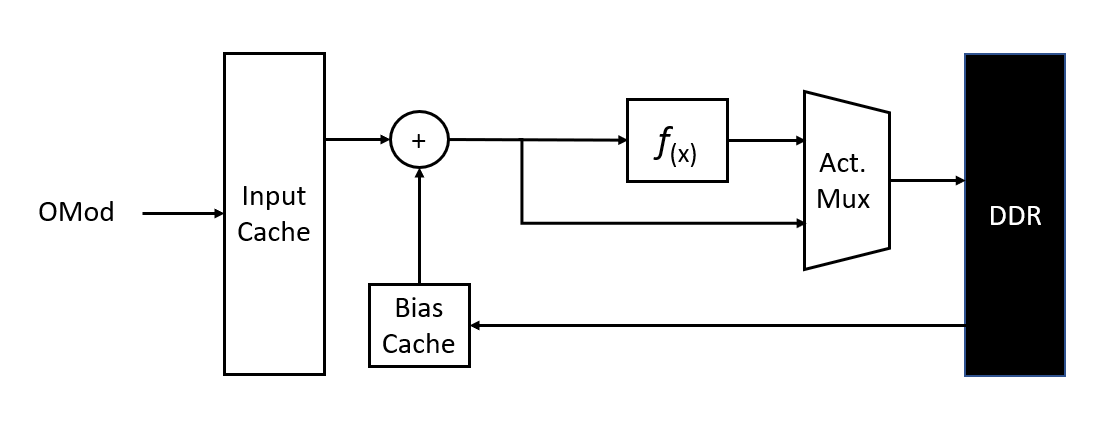} 			
	\caption{Global drain internals.}
	\label{fig:global_drain}
\end{figure}

 Once a block of data is ready to be saved back to global memory, the PEs start the draining process which begins by writing the contents of its cache to its neighbor. Results are drained in rows, so each PE drains along its column. The first row of PEs is connected to an output module (OMod) which are connected to each other in a daisy chain fashion (refer to Figure \ref{fig:systolic_array}). OMods continue the draining process by propagating the results along to a global drain whose responsibility is to prepare the data to be written back to global memory. Data being drained arrives to the output modules in a non-contiguous way, so the global drain has its own local cache that is used to buffer data back to DDR, see Figure \ref{fig:global_drain}. While this is the base design used for all our experiments, the global drain does require additional memory resources, so when scaling designs, the reordering of data may need to be done back on the host processor. Reordering data via global memory addressing is not efficient so we always write data back in a contiguous fashion. The global drain supports some additional features unique to traditional MLP style of compute. Both bias and activation function support is included and optionally bypassed if desired. In the case of bias skipping, we preload the bias cache with all zeros and never read from global memory. When bias is used, enough bias data is prefetched into a small local cache to be used on the next drain sequence. For the activation function block, we bypass simply using the activation mux as shown in Figure \ref{fig:global_drain}. 

\subsubsection{OpenCL Implementation of the 2D Systolic Array}
Each module described in the previous sections was coded in OpenCL as a separate kernel. Kernels were connected using Intel's OpenCL channels extension which implements FIFO style buffers of variable depth and width. Hooks into the modular design were made available via C99 based macro definitions. This means that at hardware compile time, the macro definitions shown in Table \ref{table:macros} need to be defined so that preprocessor can populate code according to the macros purpose. Each macro definition affects what hardware is generated, and so, we can compile various permutations of these macros to generate unique hardware for FPGA. 

\bgroup
\def\arraystretch{1.2}
\begin{table}[h]
\centering
\caption{OpenCL preprocessor macro definitions}
\begin{tabular}{|c|c|}
\hline
\textbf{Macro} & \textbf{Description}  \\ \hline
SYS\_ROWS      & Number of rows the systolic array (grid) contains. \\ \hline
SYS\_COLS      & Number of columns the grid contains                    \\ \hline
SYS\_VEC       & Width of the data path and dot product reduction tree  \\ \hline
INTERLEAVE     & Matrix A block height and matrix B block width \\ \hline
SCALE          & Number of vectors in a block \\ \hline
\end{tabular}
\label{table:macros}
\end{table}
\egroup

\begin{figure}
	\center
	\includegraphics[scale=0.40]{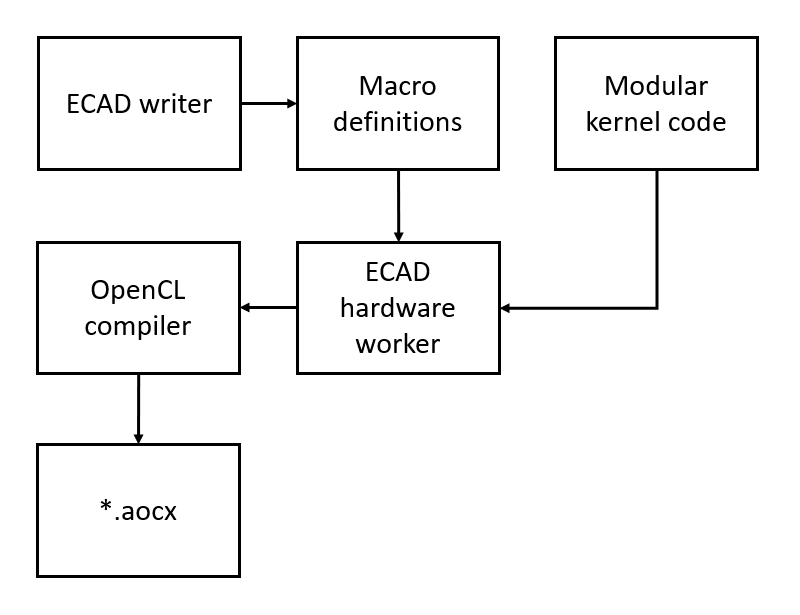} 			
	\caption{OpenCL flow for generating hardware.}
	\label{fig:opencl_flow}
\end{figure}

Figure \ref{fig:opencl_flow} shows the process of generating unique systolic array hardware. ECAD uses an actualizer, or writer, to update the macro definitions from Table \ref{table:macros}. These macros along with the OpenCL kernel code are organized by the hardware worker who runs the OpenCL compiler. The compiler goes through many steps which are detailed in \cite{IntelOpenCLProgrammingGuide} resulting in an aocx file which includes the bitstream to be programmed onto the FPGA. 

\par Kernels and elements from Figure \ref{fig:systolic_array} were connected through Intel's OpenCL channels extension which implements FIFO style buffers of variable depth and width. The data path flowing through the systolic array (not including the OMod and Global Drain connections) was designed to carry a \textit{SYS\_VEC} width that was accomplished through C99 style structs with a member array of \textit{SYS\_VEC} elements. The data path for the output modules and global drain was made to support a single element. 

\par Both loader kernels operate in a similar way except that the loader for matrix A includes additional code to handle the instructions for draining and flushing. Further, having two separate loader kernels allows for two parallel accesses to global memory. In our experiments, we only had 1 bank of DDR memory, however, for cases that have two or more, loader kernels can be made to each access a single unique bank. This helps alleviate any potential issues with bank arbitration. 

\par
\begin{figure}
	\center
	\includegraphics[scale=0.8, frame]{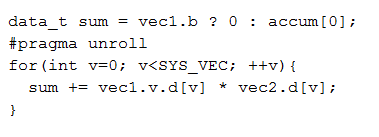}		
	\caption{OpenCL reduction tree dot product with accumulator.}
	\label{fig:dot_product}
\end{figure}

 Two different OpenCL kernels were designed for the MMods, one for the \textit{SYS\_ROWS} dimension and one for the \textit{SYS\_COLS} dimension. Like the loader kernels, the MMods that read in matrix A data needed additional logic to handle the drain and flush sequences. Figure \ref{fig:dot_product} shows the variable \textit{vec1} as a struct with member variables \textit{b (bool)} and \textit{v (vector)}. This data flows from the \textit{SYS\_ROWS} MMods who populates the bool to signal when a new set of blocks begin the next output sequence. Notice that \textit{vec2} which flows from the MMods along the \textit{SYS\_COLS} dimension only carries a vector of data \textit{d} and does not contain any additional member variables. 

\par Following the suggested method for implementing a modular systolic array as described in the Intel FPGA SDK for OpenCL Pro Edition: Programming Guide \cite{IntelOpenCLProgrammingGuide}, we leveraged the \textit{num\_compute\_units(X, Y)} attribute with compile time macro definitions \textit{SYS\_ROWS} and \textit{SYS\_COLS} to effectively stamp out a grid of processing elements. We also followed the suggestions in \cite{IntelOpenCLProgrammingGuide} for the dot product reduction tree shown in Figure \ref{fig:dot_product}. Sum is initialized with zero or the next value from the local accumulator shift register (refer to section \ref{pe}) then runs through \textit{SYS\_VEC} pipelined DSP blocks being accumulated along the way to arrive at the final reduced output. Writing the code this way helps the compiler to infer the correct Arria 10 floating point mode DSP block \cite{Arria10FloatingPointDSP}. 

\subsection{Hardware Model}
Modeling the hardware design allows the framework to search for both constrained and unconstrained designs. Unconstrained means that the evolutionary algorithm has less knowledge of the target hardware and has the freedom to search a much larger design space. When constrained, the evolutionary algorithm will only return configurations that it believes can be synthesized. Further, having a software-based model provides a much faster means of exploration compared to running through a series of hardware compiles. 

\par The framework uses a hardware worker object that is targeted for a specific design like the 2D systolic array. Neural network description files containing information about cell parameters and connectivity are sent to the worker whose job is to return the fitness of that description in context to the hardware accelerator. If the targeted hardware model can 1. provide the necessary hooks to the evolutionary algorithm to allow new permutations and 2. provide the required fitness metrics, then any accelerator can be used in the search. ECAD currently uses the 2D systolic array hardware design as the sole model that is used in search. 

\par Inputs to the current model are explained in section \ref{2D Systolic Array Implementation}. Any model that is used in ECAD must have inputs or ``hooks'' that allow the evolutionary algorithm to permute the design. After a permutation is created, it is then evaluated by the hardware worker who then provides the following results: 

\label{hwdbworker_results}
\begin{itemize}
	\item \textbf{Total time (ms)} the total time in milliseconds that it takes the accelerator to run the provided network description file.
	\item \textbf{Potential giga-operations per second (GOP/s)} the maximum performance that can be expected out of the accelerator, also known as the roofline performance.  
	\item \textbf{Effective GOP/s} actual performance of the accelerator. This number is derived by mapping the network description to the potential performance of the accelerator.
	\item \textbf{Images per second (img/s)} useful for workloads that contain image data sets such as the MNIST dataset used in most of our experiments. 
	\item \textbf{Latency (ms)} the amount of time in milliseconds before the accelerator provides the first output. 
\end{itemize}

Each result is sent back to the framework and depending on the optimization settings and various pressures used in the search, the results are evaluated and weighted to compute the overall fitness of that permutation.

\section{Experiments}

In the following experiments we use the MNIST \cite{MNISTDB} dataset to test and validate the complete ECAD framework flow. The dataset consists of 60,000 training samples and 10,000 test samples and has been used extensively in literature to demonstrate the ability of various neural network designs. Each sample is a 28x28 grayscale (1 channel) image of a single digit, which each pixel being a 8bit value ranging from 0 to 255. Before inputting a sample into the dense neural network the pixel values are compressed to a range from 0 to 1 and the sample is flattened to a single array of 784 values.
To expedite the training process, these formatted samples are passed in batches to the neural network.  

\subsection{Verifying the Hardware Model}
First, we validated our hardware model to ensure the results from our HWDBWorker were accurate. Targeting Intel's Arria 10 GX 1150 FPGA platform, the model considered various hardware resources including available DDR blocks, DSPs, and embedded memories. Averaging across several hardware compiles, we arrived at a target clock frequency of 250 MHz. Verifying the hardware model was accomplished by comparing the HWDBWorker outputs to the measured results. When describing a particular configuration for the 2D systolic array, we will use the following notation: (rows, cols, vector width, interleave, scale) which are all defined in section \ref{2D Systolic Array Implementation}. 

\par Our goal was to validate the hardware model running an actual workload to not only verify performance but also the accuracy of the results. For this, we trained a 4-layer MLP with (196/190/150/10) neurons on the MNIST dataset and compared the resulting classification accuracy with the FPGA after running all 10,000 images. The first hardware design we ran had the configuration (4, 4, 8, 8, 8). First, we created a configuration file with the hardware specifics and ran it through the HWDBWorker to determine the modeled performance of the accelerator. Next, we compiled the design and measured the hardware performance, both results are presented in Table \ref{model_vs_meas}. The resulting classification accuracy obtained from running the MLP in hardware matched the accuracy reported by the Simulator. The results show that for higher batching (64 and above), our model averages 95\% accurate and 81.7\% accurate across all tests. Lower batching for this design, which can be interpreted as a less efficient workload mapping, has a greater effect on model accuracy especially at batch 1. The smaller the workload, the shorter time the processing occurs inside the DSP blocks and the performance becomes harder to predict due to various overheads not accounted for. In other words, the variance we see in the measured results becomes less significant with larger workloads. The model is always initialized with the theoretical ceiling for performance and works its way down once it considers the workload. Two observations from our results are: 1. This model is providing the theoretical limit for these workloads so instead of looking at the model and attempting to change it to match the inefficiencies of the hardware, we will instead look at optimizing the hardware further to match the model and 2. our model remains consistent across permutations so that while some absolute performance results may vary (low batch for example) the relationship between permutations will always remain the same.

\bgroup
\def\arraystretch{1.2}
\begin{table}[h]
\centering
\caption{Modeled vs measured performance for configuration (4, 4, 8, 8, 8) running an MLP}
\begin{tabular}{c|c|c|c|c|}
\cline{2-5}
\multicolumn{1}{l|}{}                     & \multicolumn{2}{c|}{\textbf{Modeled}}                   & \multicolumn{2}{c|}{\textbf{Measured}}                  \\ \hline
\multicolumn{1}{|c|}{\textbf{Batch size}} & \textbf{Effective GOP/s} & \textbf{Execution time (ms)} & \textbf{Effective GOP/s} & \textbf{Execution time (ms)} \\ \hline
\multicolumn{1}{|c|}{1}                   & 1.16                     & 0.38                         & 0.65                     & 0.679                        \\ \hline
\multicolumn{1}{|c|}{16}                  & 18.6                     & 0.38                         & 10.3                     & 0.68                         \\ \hline
\multicolumn{1}{|c|}{32}                  & 37.2                     & 0.38                         & 20.7                     & 0.68                         \\ \hline
\multicolumn{1}{|c|}{64}                  & 40.3                     & 0.7                          & 35.81                    & 0.78                         \\ \hline
\multicolumn{1}{|c|}{128}                 & 42                       & 1.35                         & 39                       & 1.43                         \\ \hline
\multicolumn{1}{|c|}{256}                 & 42.98                    & 2.63                         & 41.2                     & 2.74                         \\ \hline
\multicolumn{1}{|c|}{512}                 & 43.47                    & 5.2                          & 41.5                     & 5.45                         \\ \hline
\multicolumn{1}{|c|}{1024}                & 43.7                     & 10.35                        & 42.4                     & 10.66                        \\ \hline
\multicolumn{1}{|c|}{2048}                & 43.84                    & 20.64                        & 43                       & 21                           \\ \hline
\end{tabular}
\label{model_vs_meas}
\end{table}
\egroup

\subsection{Design Space Exploration}
In the following design experiments, our goal is to find an optimized hardware solution for a given network structure that consists of an input layer, a single hidden layer and an output layer (784/N/10). The input is a 28X28 image and the output is the digit found in the input image.

We start investigating the different individual pressures on the design by running an evolutionary process on individual goals.

\begin{figure}
	\center
	\includegraphics[scale=0.50]{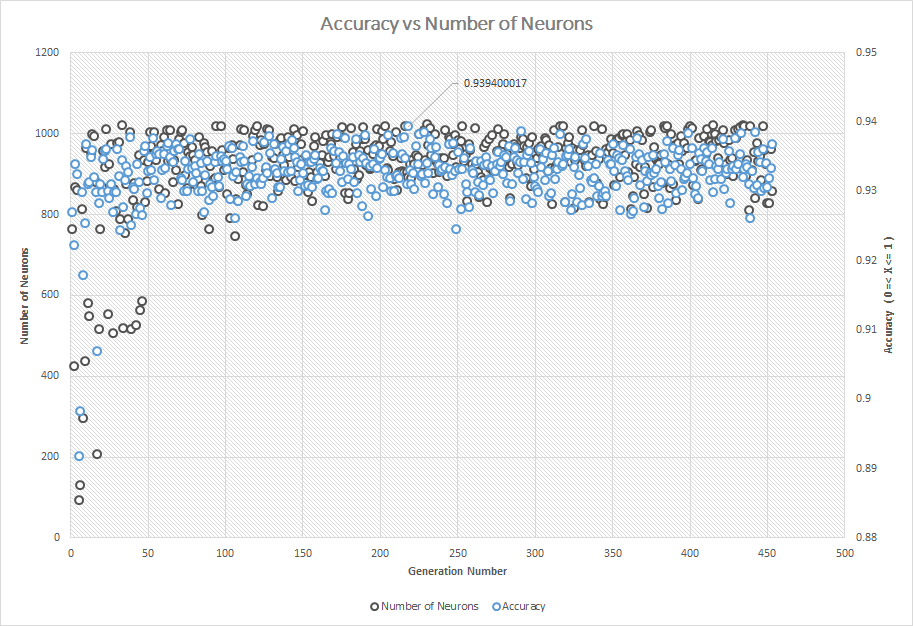} 			
	\caption{Optimize for accuracy - accuracy vs number of neurons }
	\label{fig:accuracy_vs_neurons}	
\end{figure}

\begin{figure}
	\center
	\includegraphics[scale=0.50]{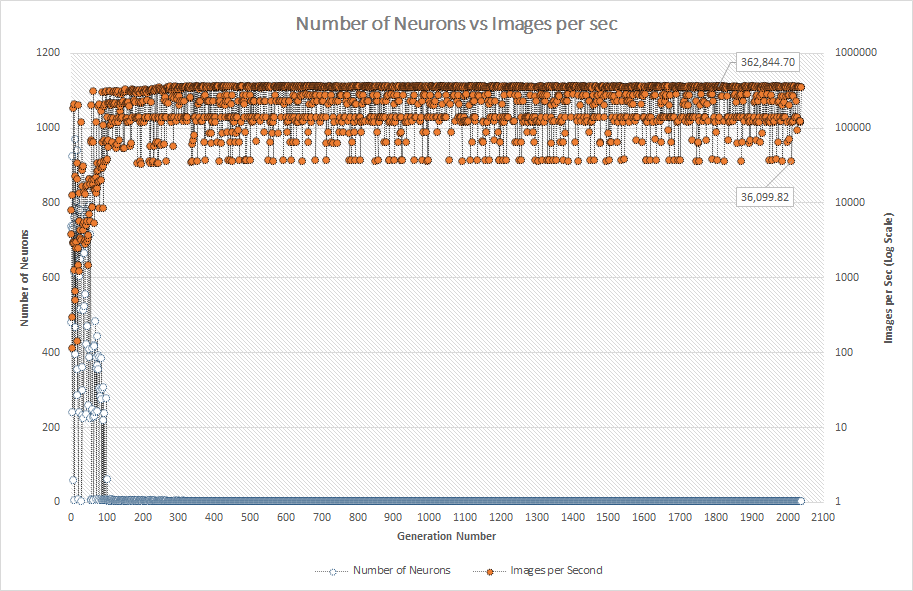} 			
	\caption{Optimize for img/s - images per sec vs number of neurons }
	\label{fig:images_per_sec_vs_neurons}	
\end{figure}

\begin{figure}
	\center
	\includegraphics[scale=0.50]{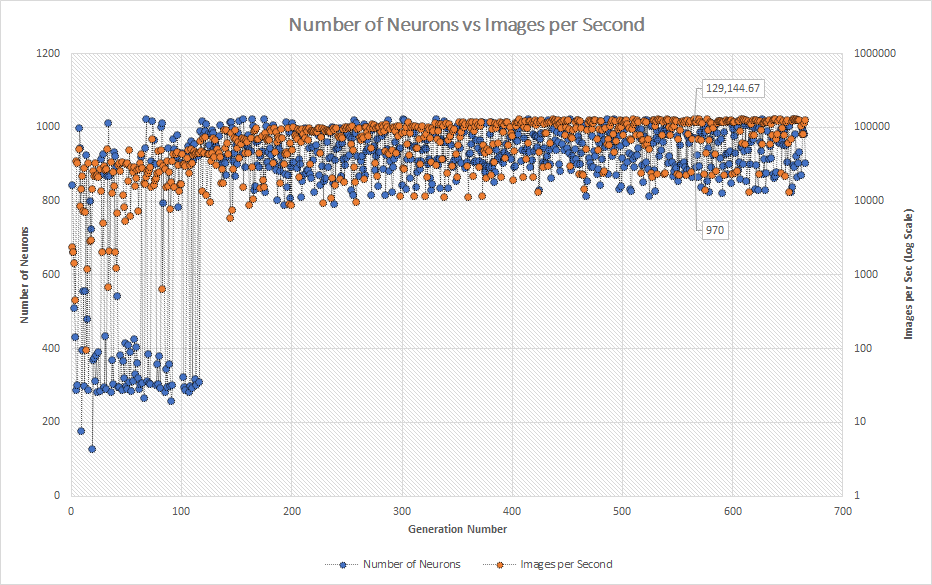} 			
	\caption{Optimize for img/s and min accuracy - images per sec vs number of neurons }
	\label{fig:images_per_sec_vs_neurons_opt_both}	
\end{figure}

\begin{figure}
	\center
	\includegraphics[scale=0.45]{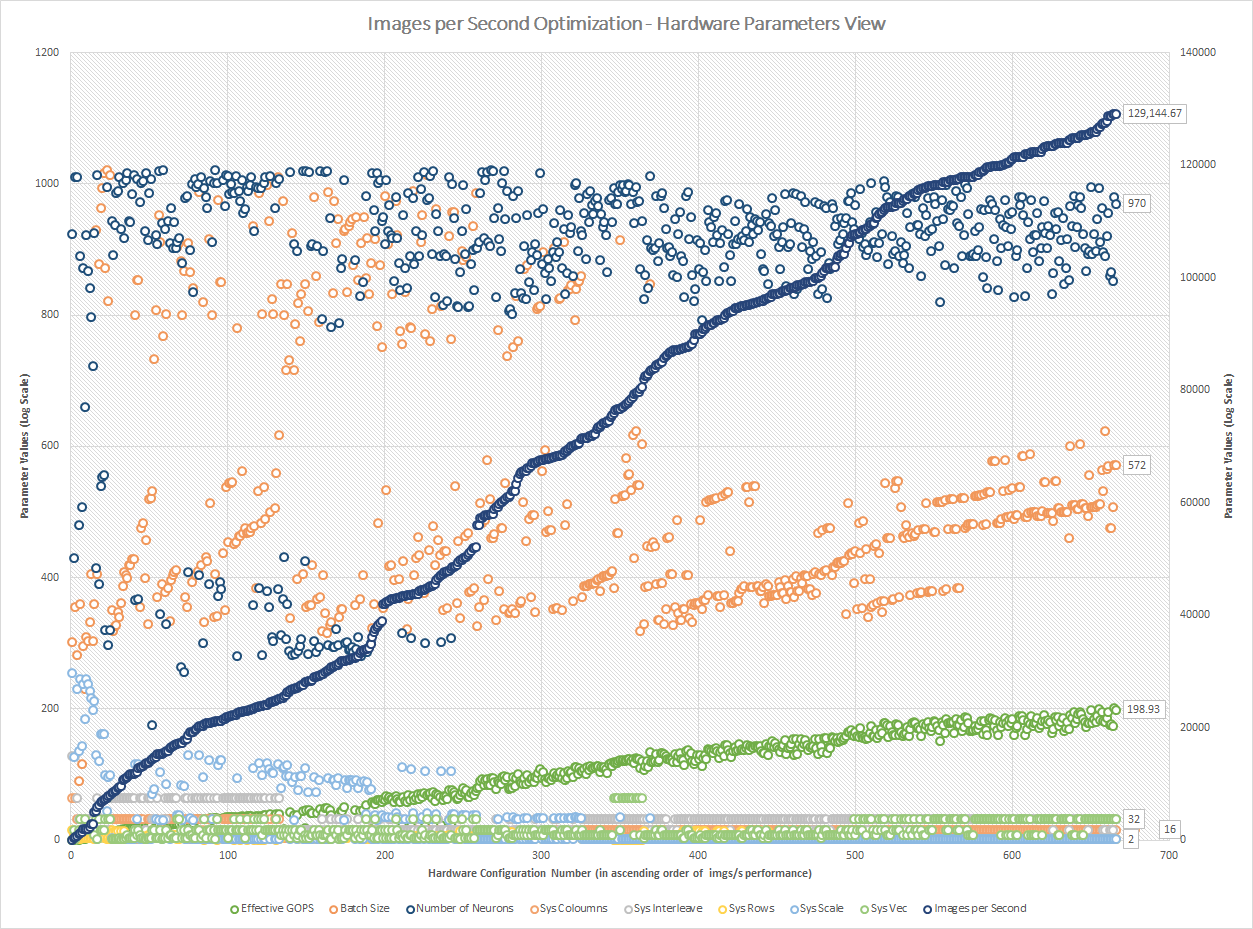} 			
	\caption{Optimize for img/s and min accuracy - convergence of hardware parameters from least performant hardware architecture to most performant architecture when optimizing for img/s objective }
	\label{fig:images_per_sec_vs_neurons_opt_both_hw_params}	
\end{figure}

\subsubsection{Optimizing for Accuracy }

To optimize for accuracy, we select to run the simulator worker and create an evolutionary pressure towards accuracy. We do this by setting the goal to maximize accuracy. The simulator will be given different network designs, it will then simulate them using TensorFlow for a given number of epochs (4) and return the accuracy results. Figure \ref{fig:accuracy_vs_neurons} shows accuracy vs number of neurons as the evolutionary process progresses through the generations. As can be seen in Figure \ref{fig:accuracy_vs_neurons} after 50 generations the number of neurons rises to above 700. As the evolutionary process progresses it finds that to achieve high accuracy the number of neurons must fluctuate to between 748 and 1024 with the best result at 894. Some support for these results can be found in the literature. Several prior works report competitive results for a single hidden layer of neurons containing 800 neurons \cite{MNISTDB}.

\subsubsection{Optimizing for Images per Second}
In this stage our goal is to try to force the evolving hardware design to output the most images per second. This would be a common goal in hardware design exploration especially when using re-configurable hardware with a low overall power envelope. By maximizing the number of images per second (img/s) we could potentially save money and conserve energy.
To create such evolutionary pressure, we use our HWDBWorker to evaluate potential designs and return information such as the images per second (see \ref{hwdbworker_results}) which we could use to select upon in the evolutionary process. 
As can be seen in Figure \ref{fig:images_per_sec_vs_neurons}, to achieve an optimal number of img/s, the evolutionary process put pressure on the number of neurons to drop. In fact, to achieve the optimal performance, the number of neurons tends towards zero, this is because the EA found that less computation, i.e., less neurons, resulted in a higher throughput and because there was no pressure to maintain accuracy, it continued to drop. During this particular search, the hardware configuration started to converge (4,8,8,16,18) while the batch size began to rise (to 1024) and the neurons remained low. This resulted in hardware that was 100\% efficient in the ``batch'' dimension which is used in every layer. Since the MNIST dataset requires an input size of 784, the EA needed to find a common dimension through the vector width and scale parameters that both kept total execution time of the MLP down but could also process the input efficiently. This dimension mapped to 91\% efficiency. The trend of reducing the neurons did work for the EA but had pressure been placed on accuracy, it would have found it could utilize the entire common dimension with neurons.

\subsection{Optimizing for both Accuracy and Images per Second}
Now that we understand the different conflicting pressures we proceed to try and find an optimal compromise between the accuracy and the images per second (img/s) metric. If we do not give a constraint on each of the objectives our evolutionary process will find some middle ground compromise which might not be suited for our needs. So instead we force the evolutionary process towards a minimum accuracy goal and a maximum number of img/s goal. 
Figure \ref{fig:images_per_sec_vs_neurons_opt_both} shows the resulting evolutionary process and the optimal combination of img/s. The minimum accuracy goal is set to ninety percent. The best img/s result (129,144) is lower than the unconstrained accuracy version (362,844) but given a minimum accuracy of 90\% this result could be used in a practical design.

Trends in the hardware configurations that the EA displayed can be visualized in Figure \ref{fig:images_per_sec_vs_neurons_opt_both_hw_params}. We noted previously that to get the best accuracy, the algorithm uses more neurons, yet to get the best performance, typically less neurons are desired. This forces the EA to find better results for how well the neurons are mapping on hardware. Table \ref{tbl_acc_imgs} shows the results of the top permutations for the optimization process. The only parameters that seem to be different are the columns, vector width, and interleaving factor. Rows seem to have converged to 2 because input to an MLP without batching is a vector and the EA decided that making this dimension in hardware small, better fits the solution. Further, batching during inference does not affect accuracy so the hardware is better utilized in other dimensions. Scale also converges at 2 and this is most likely because this parameters can be thought of as quantization along the matrix common dimensions, and a smaller number may result in a better fit. Beyond efficiency mapping, scale also provides a trade-off with global (off-chip) memory bandwidth, and because our model considers DDR memory reads and writes, scale will generally tend to be lower to help balance the transactions. The third column, fitness sum, gives the results of what the EA found to be the best fit solution given pressures. We notice that all the results lie in between the top permutations for accuracy and effective GOP/s. These results show how small variations in permutations yield large variations in fitness evaluations, and in most cases, top performers arrive at unique solutions that may never be considered if designed by hand. 

\bgroup
\def\arraystretch{1.2}
\begin{table}[h]
\centering
\caption{Top permutations from optimizing accuracy and img/s}
\begin{tabular}{r|c|c|c|}
\cline{2-4}
\multicolumn{1}{l|}{}                          & \multicolumn{3}{c|}{\textbf{Top permutations}}                      \\ \cline{2-4} 
\multicolumn{1}{l|}{}                          & \textbf{Accuracy} & \textbf{Effective GOP/s} & \textbf{Fitness Sum} \\ \hline
\multicolumn{1}{|r|}{\textbf{Accuracy}}        & 0.942             & 0.933                    & 0.936                \\ \hline
\multicolumn{1}{|r|}{\textbf{HW Config}}       & 2,8,16,16,2       & 2,16,32,32,2             & 2,8,32,16,2          \\ \hline
\multicolumn{1}{|r|}{\textbf{Effective GOP/s}} & 92.62             & 200.98                   & 174                  \\ \hline
\multicolumn{1}{|r|}{\textbf{Batch size}}      & 484               & 572                      & 508                  \\ \hline
\multicolumn{1}{|r|}{\textbf{Neurons}}         & 1018              & 980                      & 852                  \\ \hline
\end{tabular}
\label{tbl_acc_imgs}
\end{table}
\egroup

\subsection{Hardware Generation and Measured Results}
Now that we have a list of feasible permutations with the desired optimizations, we use the ECAD hardware flow to synthesis the design. This is achieved by exporting the neural network from the ECAD database into a network description file that can be fed to the ECAD framework for partial or complete hardware compile. A simple command line accepts the exported design and starts the build process. 

\par During search, the HWDBWorker only returns configurations it believes to be valid, however, there are some cases where a design may not pass hardware compilation due to excess resource utilization. The recommended next step after search completes is to run a partial compile on the top performers. Partial compiles provide detailed information on resource utilization but the process takes a few minutes to complete which is why partial compiles are done after search completes. After finding the top performer that passes the partial compilation stage, it is sent through the full hardware flow which includes bitstream generation and testing in hardware. Currently, this process of weeding out designs that cannot produce hardware is done by hand, but could be automated into a post search process. 

\par Performance results from running the top permutations in Table \ref{tbl_acc_imgs} through hardware are presented in Table \ref{hardware_perf_results} and their corresponding logic utilization are presented in Table \ref{hardware_util_results}. All the top permutations in Table \ref{tbl_acc_imgs} were valid configurations and passed the full hardware compilation stage. For convenience, the hardware configuration notation is rows, columns, vector width, interleaving, and scale. The execution time measured in milliseconds is the total time it takes to run a batch of MNIST images through the 2-layer MLP in hardware. These results further validate the hardware model. Table \ref{hardware_util_results} shows as a percentage, how many resources each design is using. Note that each Fmax is slightly lower than the 250 MHz we searched at. This, in part, is due to only compiling a single seed design and having run more compiles could have resulted in an Fmax closer to that target. The model execution time listed in the table was scaled from 250 MHz to the actual MHz that the design closed at.

\bgroup
\def\arraystretch{1.2}
\begin{table}[h]
\centering
\caption{Hardware performance results compared to model for top permutations}
\begin{tabular}{|c|c|c|}
\hline
\textbf{Hardware Configuration} & \textbf{Modeled Execution Time (ms)} & \textbf{Measured Execution Time (ms)} \\ \hline
2,8,16,16,2                     & 9.59                                 & 9.64                                  \\ \hline
4,8,8,16,18                     & 4.66                                 & 4.65                                  \\ \hline
2,8,32,16,2                     & 4.64                                 & 5.53                                  \\ \hline
\end{tabular}
\label{hardware_perf_results}
\end{table}
\egroup

\bgroup
\def\arraystretch{1.2}
\begin{table}[h]
\centering
\caption{FPGA resource utilization for top permutations}
\begin{tabular}{c|c|c|c|c|c|c|c|}
\cline{2-8}
\multicolumn{1}{l|}{}                                 & \multicolumn{4}{c|}{\textbf{Hardware Compile}}              & \multicolumn{3}{c|}{\textbf{Partial Compile}}                                                              \\ \hline
\multicolumn{1}{|c|}{\textbf{Hardware Configuration}} & \textbf{Fmax} & \textbf{ALM} & \textbf{M20K} & \textbf{DSP} & \multicolumn{1}{l|}{\textbf{ALM}} & \multicolumn{1}{l|}{\textbf{M20K}} & \multicolumn{1}{l|}{\textbf{DSP}} \\ \hline
\multicolumn{1}{|c|}{2,8,16,16,2}                     & 220.41        & 37\%         & 36\%          & 26\%         & 47\%                              & 38\%                               & 26\%                              \\ \hline
\multicolumn{1}{|c|}{4,8,8,16,18}                     & 228.26        & 38\%         & 34\%          & 26\%         & 45\%                              & 37\%                               & 26\%                              \\ \hline
\multicolumn{1}{|c|}{2,8,32,16,2}                     & 212.18        & 50\%         & 30\%          & 43\%         & 61\%                              & 55\%                               & 43\%                              \\ \hline
\end{tabular}
\label{hardware_util_results}
\end{table}
\egroup

\section{Conclusions}
We present a novel software framework called Evolutionary Cell Aided Design(ECAD) that aids in the exploration and design of efficient Neural Network Architectures(NNAs) for reconfigurable hardware. Given a general neural network structure and a set of constraints and fitness functions, ECAD uses a reconfigurable-hardware/NNA co-design approach to optimize designs on both the NNA and the hardware side and attempts to find the fittest solutions according to a predefined set of goals. We discuss and demonstrate the ability to optimize for different and multiple optimization objectives. Performance data of the top permutations for different optimizations including accuracy, img/s, and both simultaneously, is presented along with the complete end-to-end ECAD flow targeting an Intel FPGA Arria 10 1150 GX device. Finally, comparing the hardware performance of the top permutations against our model served to validate the results of the evolutionary process.

\section{Acknowledgments}
Limited portion of this work has previously been presented at \textit{The 1st International Workshop on FPGAs for Domain Experts(FPODE-18)\cite{FPODE18}}.

\printbibliography[ heading=bibintoc, title={References}]

\clearpage
\begin{appendices}

\section{Sample Configuration File}

\begin{lstlisting}[language=json,firstnumber=1, caption = {ECAD configuration file sample}, label={lst:ecad_cfg_lst}]
{
"name": "MLP Example configuration file",
"comment": "Test cell array for MLP",
"includes_comment": "list of configuration include files",
"includes": [ "GlobalSettings.ecad.cfg" ],
"version": "v0.0.3b",

"popConfigValues":  
{
  "comment": "population configuration values - these get loaded into the cell network population class in the evolution engine",
  "initialPopSize": 20,
  "maxPopSize": 40,
  "changeRate": 0.20,
  "minIndivEvalCompleteBeforeFitSelect": 10,
  "maxGenerations": 2000,
  "fitnessScoreGoal": 2.0,
  "evalTypes":
  [
    { "type": "simJob",  "weight": 1.0, "minValue": 0.9, "maxValue": 1, "active": true, "allowOverflow": false, "epochs": 4, "batchSize": 100},
    { "type": "hwDBJob", "weight": 1.0, "minValue": 0, "maxValue": 1000.0, "active": true, "allowOverflow": false  },
    { "type": "physJob", "weight": 1.0, "minValue": 0, "maxValue": 1, "active": false,  "allowOverflow": false, "minimize": true}
  ]
},
"traitConfigValues":
{
  "defChangeRate": 0.10
},

"cellConfigValues":  
[
],

"cellTypes": 
[ 
  {
    "comment"     : "Input Layer",
    "cell_type"   : "input",
    "batch_size"  : { "minValue": 2, "maxValue": 1024, "modValue": 2},

    "templateFile": "",
    "mainFuncName": ""
  },
  {
    "comment"   : "Dense Layer",
    "cell_type" : "dense",

    "systolic_id" : 0,

    "neurons"   : { "minValue": 2, "maxValue": 1024, "modValue": 2, "changeRate": 0.1 }, 
    "sys_rows"  : { "minValue": 2, "maxValue": 64,  "modValue": 2, "changeRate": 0.1 }, 
    "sys_cols"  : { "minValue": 2, "maxValue": 64,  "changeRate": 0.5, "powValue": 2, "func": "PowFunction"}, 
    "sys_vec"   : { "minValue": 2, "maxValue": 64,  "changeRate": 0.5, "powValue": 2, "func": "PowFunction"}, 

    "sys_intrlv-comment" : "DenseCell mutate sets sys_intrlv = (random pow 2 >= sys_rows+sys_cols)", 
    "sys_intrlv" : { "minValue": 2, "maxValue": 256, "modValue": 2},
    "sys_scale"  : { "minValue": 2, "maxValue": 256, "modValue": 2},

    "row_blocks" : 0,
    "col_blocks" : 0,
    "vec_blocks" : 0,
    "arows_pad"  : 0,
    "acols_pad"  : 0,
    "bcols_pad"  : 0,

    "enableBias" : { "minValue": 0, "maxValue": 1},

    "HWGenMode_comment": "Hardware Generation Mode - for now options are: Single Systolic Array(SSA), Multi Systolic Array(MSA)",
    "HWGenMode": "SSA",

    "weightsManipulation": "matmul", 
    "layerManipulation": "add",

    "templateFile"  : "dense_cell.h",
    "mainFuncName"  : "dense_cell"
  },
  {
    "comment"   : "Relu Layer",
    "cell_type" : "relu",
    "relu_vec"  : 1,
    "templateFile"  : "relu_cell.h",
    "mainFuncName"  : "relu_cell"
 },
 {
    "comment"     : "Output Layer",
    "cell_type"   : "output",

    "templateFile": "",
    "mainFuncName": ""
  }
],

"netConfig":  
{
  "netType"      : "mlp",
  "templateFile" : "",
  "mainFuncName" : ""
},

"hwConfig":
{
  "comment" : "These are the resource limits provided by the device data sheet", 
  "deviceType": "Arria10-1150",
  "dsp": 1518,     
  "freq": 250,
  "sram": 54260,
  "mem_banks": 1,
  "mem_speed": 2400, 
  "mem_rate": 8 
},

"cellArray":
[
  {"comment": "input layer", "cell_type": "input",  "cell_name": "X",       "input": "global",  "output": "dense00", "input_size": 784, "fixed": true}, 
  {"comment": "dense layer", "cell_type": "dense",  "cell_name": "dense00", "input": "X",       "output": "relu00",  "fixed": false}, 
  {"comment": "relu layer",  "cell_type": "relu",   "cell_name": "relu00",  "input": "dense00", "output": "Y",       "fixed": false}, 
  {"comment": "output layer","cell_type": "output", "cell_name": "Y",       "input": "relu00",  "output": "global",  "output_size": 10, "fixed": true} 
]
} 

\end{lstlisting}

\end{appendices}

\end{document}